\documentclass[11pt,a4paper]{article}

\pdfoutput=1

\usepackage[hyperref]{acl2017}
\usepackage{times, latexsym, amsmath, subcaption, multirow, dblfloatfix}

\usepackage[pdftex]{graphicx}

\def\aclpaperid{43} 
\aclfinalcopy 

\newcommand{\figref}[1]{Figure~\ref{fig:#1}}
\newcommand{\tabref}[1]{Table~\ref{tab:#1}}
\newcommand{\sectionref}[1]{Section~\ref{sec:#1}}

\newcommand{\rot}[1]{\rotatebox{90}{#1}}

\title{Binary Paragraph Vectors}

\author{Karol Grzegorczyk \and Marcin Kurdziel\\
AGH University of Science and Technology \\
Department of Computer Science \\
Krakow, Poland \\
\texttt{\{kgr,kurdziel\}@agh.edu.pl}}

\date{}

\begin{document}

\maketitle

\begin{abstract}
Recently Le \& Mikolov described two log-linear models, called Paragraph Vector, that can be used to learn
state-of-the-art distributed representations of documents. Inspired by this work, we
present \emph{Binary Paragraph Vector} models: simple neural networks that learn short binary codes for fast
information retrieval. We show that binary paragraph vectors outperform autoencoder-based binary
codes, despite using fewer bits. We also evaluate their precision in transfer learning settings, where
binary codes are inferred for documents unrelated to the training corpus. Results from these experiments
indicate that binary paragraph vectors can capture semantics relevant for various domain-specific documents.
Finally, we present a model that simultaneously learns short binary codes and longer, real-valued
representations. This model can be used to rapidly retrieve a short list of highly relevant documents from a
large document collection.
\end{abstract}

\section{Introduction}\label{sec:introduction}

One of the significant challenges in contemporary information processing is the sheer volume of available data.
\citet{gantz2012digital}, for example, claim that the amount of digital data in the world doubles every two years.
This trend underpins efforts to develop algorithms that can efficiently search for relevant information in huge datasets.
One class of such algorithms, represented by, e.g., Locality Sensitive Hashing~\citep{indyk1998approximate}, relies on
hashing data into short, locality-preserving binary codes~\citep{wang2014hashing}. The codes can then be used to group the
data into buckets, thereby enabling sublinear search for relevant information, or for fast comparison of data items.
Most of the algorithms from this family are data-oblivious, i.e. can generate hashes for any type of data.
Nevertheless, some methods target specific kind of input data, like text or image.

In this work we focus on learning binary codes for text documents. An important work in this direction has been presented
by~\citet{salakhutdinov2009semantic}. Their \emph{semantic hashing} leverages autoencoders with sigmoid bottleneck layer to
learn binary codes from a word-count bag-of-words~(BOW) representation. Salakhutdinov~\&~Hinton report that binary codes
allow for up to $20$-fold improvement in document ranking speed, compared to real-valued
representation of the same dimensionality. Moreover, they demonstrate that semantic hashing codes used as an initial document
filter can improve precision of
TF-IDF-based retrieval. Learning binary representation from BOW, however, has its disadvantages. First, word-count representation,
and in turn the learned codes, are not in itself stronger than TF-IDF. Second, BOW is an inefficient representation: even for
moderate-size vocabularies BOW vectors can have thousands of dimensions. Learning fully-connected autoencoders for such
high-dimensional vectors is impractical. Salakhutdinov~\&~Hinton restricted the BOW vocabulary in their experiments
to 2000 most frequent words.

Binary codes have also been applied to cross-modal retrieval where text is one of the modalities. Specifically,
\citet{wang2013semantic} incorporated tag information that often accompany text documents, while
\citet{masci2014multimodal} employed siamese neural networks to learn single binary representation for text and image data.

Recently several works explored simple neural models for unsupervised learning of distributed representations of words,
sentences and documents. \citet{mikolov2013efficient} proposed log-linear models that learn distributed representations of
words by predicting a central word from its context (CBOW model) or by predicting context words given the central
word (Skip-gram model). The CBOW model was then extended by \citet{le2014distributed} to learn distributed representations
of documents. Specifically, they proposed Paragraph Vector Distributed Memory (PV-DM) model, in which the central word is predicted
given the context words and the document vector. During training, PV-DM learns the word embeddings and the parameters of the
softmax that models the conditional probability distribution for the central words. During inference, word embeddings and softmax
weights are fixed, but the gradients are backpropagated to the inferred document vector. In addition to PV-DM, Le~\&~Mikolov
studied also a simpler model, namely Paragraph Vector Distributed Bag of Words (PV-DBOW). This model predicts words in the document
given only the document vector. It therefore disregards context surrounding the predicted word and does not learn word embeddings.
Le~\&~Mikolov demonstrated that paragraph vectors outperform BOW and bag-of-bigrams in information retrieval task, while using
only few hundreds of dimensions. These models are also amendable to learning and inference over large vocabularies. Original CBOW
network used hierarchical softmax to model the probability distribution for the central word. One can also use noise-contrastive
estimation~\citep{gutmann2010noise} or importance sampling~\citep{cho2015using} to approximate the gradients with respect to the
softmax logits.

An alternative approach to learning representation of pieces of text has been recently described by \citet{kiros2015skip}. Networks
proposed therein, inspired by the Skip-gram model, learn to predict surrounding sentences given the center sentence. To this
end, the center sentence is encoded by an encoder network and the surrounding sentences are predicted by a decoder network conditioned
on the center sentence code. Once trained, these models can encode sentences without resorting to backpropagation inference. However,
they learn representations at the sentence level but not at the document level.

In this work we present Binary Paragraph Vector models, an extensions to PV-DBOW and PV-DM that learn short binary codes for text
documents. One inspiration for binary paragraph vectors comes from a recent work by~\citet{lin2015deep} on learning binary codes for
images. Specifically, we introduce a sigmoid layer to the paragraph vector models, and train it in a way that encourages binary
activations. We demonstrate that the resultant binary paragraph vectors significantly outperform semantic hashing codes. We
also evaluate binary paragraph vectors in transfer learning settings, where training and inference are carried out on unrelated text
corpora. Finally, we study models that simultaneously learn short binary codes for document filtering and longer, real-valued
representations for ranking. While~\citet{lin2015deep} employed a supervised criterion to learn image codes, binary paragraph
vectors remain unsupervised models: they learn to predict words in documents.

\section{Binary paragraph vector models}\label{sec:models}

The basic idea in binary paragraph vector models is to introduce a sigmoid nonlinearity before the softmax that models the conditional
probability of words given the context. If we then enforce binary or near-binary activations in this nonlinearity, the probability
distribution over words will be conditioned on a bit vector context, rather than real-valued representation. The inference in
the model proceeds like in Paragraph Vector, except the document code is constructed from the sigmoid activations. After rounding,
this code can be seen as a distributed binary representation of the document.

\begin{figure}[!b]
  \centering
  \includegraphics[width=1.0\linewidth]{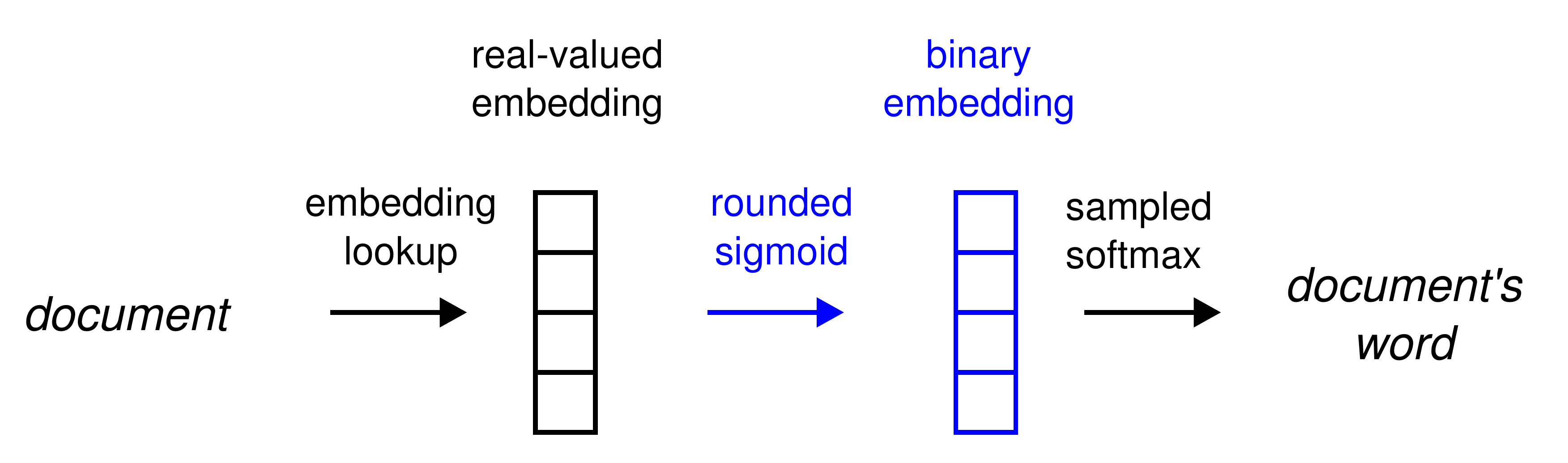}
  \caption{The Binary PV-DBOW model. Modifications to the original PV-DBOW model are highlighted.}
  \label{fig:pv-dbow-bin}
\end{figure}

\begin{figure}[htb!]
  \centering
  \includegraphics[width=1.0\linewidth]{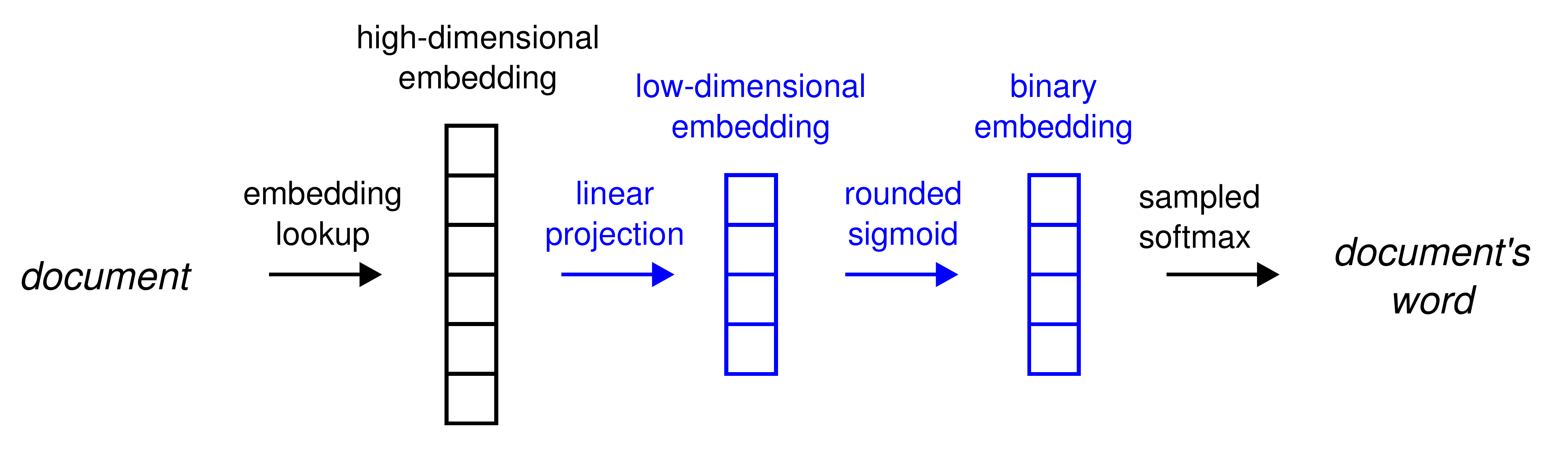}
  \caption{The Real-Binary PV-DBOW model. Modifications to the original PV-DBOW model are highlighted.}
  \label{fig:pv-dbow-bin-projections}
\end{figure}

In the simplest Binary PV-DBOW model~(\figref{pv-dbow-bin}) the dimensionality of the real-valued document embeddings is equal to the
length of the binary codes. Despite this low dimensional representation~--~a~useful binary hash will typically have 128 or fewer
bits~--~this model performed surprisingly well in our experiments.
Note that we cannot simply increase the embedding dimensionality in Binary PV-DBOW in order to learn better codes: binary vectors
learned in this way would be too long to be useful in document hashing. The retrieval performance can, however, be improved by using
binary codes for initial filtering of documents, and then using a representation with higher capacity to rank the remaining documents
by their similarity to the query. \citet{salakhutdinov2009semantic}, for example, used semantic hashing codes for initial filtering
and TF-IDF for ranking. A similar document retrieval strategy can be realized with binary paragraph vectors. Furthermore, we
can extend the Binary PV-DBOW model to simultaneously learn short binary codes and higher-dimensional real-valued representations.
Specifically, in the Real-Binary PV-DBOW model~(\figref{pv-dbow-bin-projections}) we introduce a linear projection between the document
embedding matrix and the sigmoid nonlinearity. During training, we learn the softmax parameters and the projection matrix. During
inference, softmax weights and the projection matrix are fixed. This way, we simultaneously obtain a high-capacity representation of a document in
the embedding matrix, e.g. \mbox{$300$-dimensional} real-valued vector, and a short binary representation from the sigmoid activations.
One advantage of using the Real-Binary PV-DBOW model over two separate networks is that we need to store only one set of softmax
parameters (and a small projection matrix) in the memory, instead of two large weight matrices. Additionally, only one model needs to be
trained, rather than two distinct networks.

\begin{figure}[!t]
  \centering
  \includegraphics[width=1.0\linewidth]{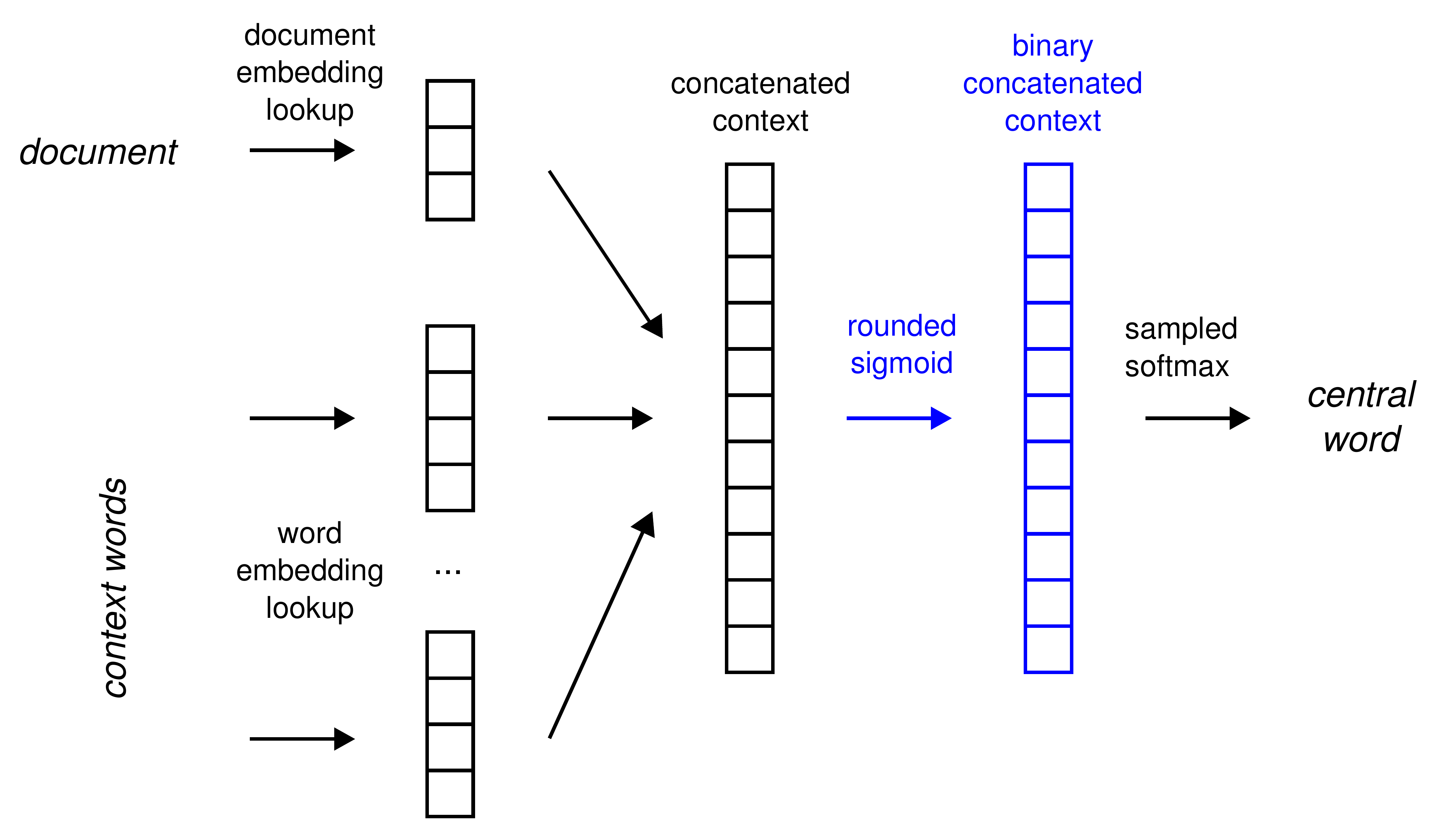}
  \caption{The Binary PV-DM model. Modifications to the original PV-DM model are highlighted.}
  \label{fig:pv-dm-bin}
\end{figure}

Binary document codes can also be learned by extending distributed memory models. \citet{le2014distributed} suggest that in PV-DM, a context
of the central word can be constructed by either concatenating or averaging the document vector and the embeddings of the surrounding words.
However, in Binary PV-DM~(\figref{pv-dm-bin}) we always construct the context by concatenating the relevant vectors before applying the
sigmoid nonlinearity. This way, the length of binary codes is not tied to the dimensionality of word embeddings.

Softmax layers in the models described above should be trained to predict words in documents given binary context vectors. Training
should therefore encourage binary activations in the preceding sigmoid layers. This can be done in several ways. In semantic hashing
autoencoders~\citet{salakhutdinov2009semantic} added noise to the sigmoid coding layer. Error backpropagation then countered the noise, by
forcing the activations to be close to 0 or 1. Another approach was used by~\citet{krizhevsky2011using} in autoencoders that learned
binary codes for small images. During the forward pass, activations in the coding layer were rounded to 0 or 1. Original (i.e. not rounded)
activations were used when backpropagating errors. Alternatively, one could model the document codes with stochastic binary neurons. Learning
in this case can still proceed with error backpropagation, provided that a suitable gradient estimator is used alongside stochastic activations.
We experimented with the methods used in semantic hashing and Krizhevsky's autoencoders, as well as with the two biased gradient estimators
for stochastic binary neurons discussed by~\citet{bengio2013estimating}. We also investigated the slope annealing
trick~\citep{chung2016hierarchical} when training networks with stochastic binary activations. From our experience, binary paragraph vector
models with rounded activations are easy to train and learn better codes than models with noise-based binarization or stochastic neurons. We
therefore use Krizhevsky's binarization in our models.

\section{Experiments}\label{sec:experiments}

To assess the performance of binary paragraph vectors, we carried out experiments on three datasets: 20 Newsgroups\footnote{Available at
\url{http://qwone.com/~jason/20Newsgroups}}, a cleansed version (also called v2) of Reuters Corpus Volume~1\footnote{Available at
\url{http://trec.nist.gov/data/reuters/reuters.html}}~(RCV1) and English Wikipedia\footnote{A snapshot from April 5th, 2016}. As paragraph vectors can
be trained with relatively large vocabularies, we did not perform any stemming of the source text. However, we removed stop words as well as
words shorter than two characters and longer than 15 characters. Results reported by~\cite{li2015learning} indicate that performance\
of PV-DBOW can be improved by including \emph{n-grams} in the model. We therefore evaluated two variants of Binary PV-DBOW: one
predicting words in documents and one predicting words and bigrams. Since 20 Newsgroups is a relatively small dataset, we used all
words and bigrams from its documents. This amounts to a vocabulary with slightly over one million elements. For the RCV1 dataset we used
words and bigrams with at least 10 occurrences in the text, which gives a vocabulary with approximately 800 thousands elements.
In case of English Wikipedia we used words and bigrams with at least 100 occurrences, which gives a vocabulary with
approximately 1.5 million elements.

\begin{figure*}[htb!]
  \begin{tabular}{cc}
    \multicolumn{2}{c}{(a) 20 Newsgroups} \\
    \includegraphics[width=0.48\textwidth]{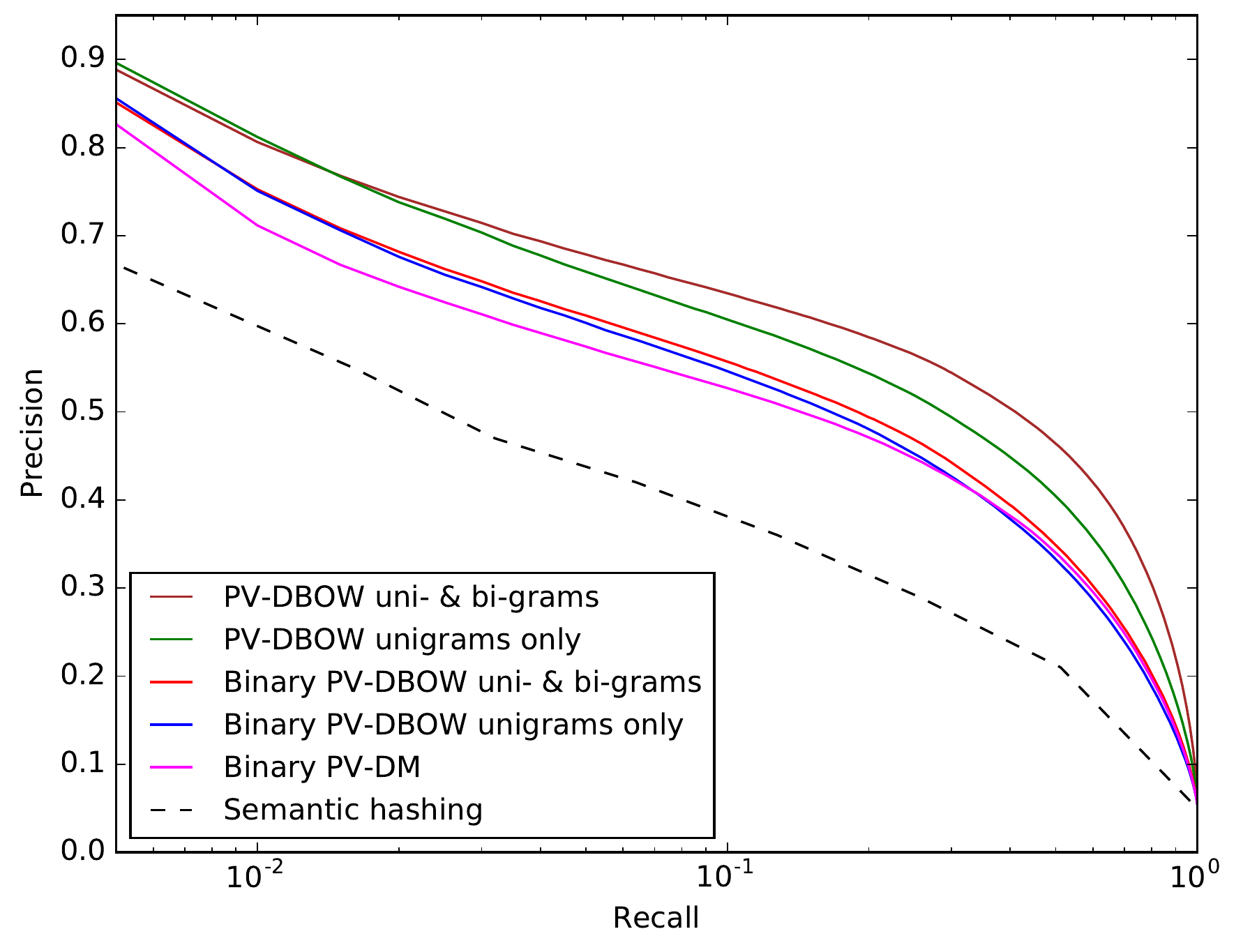} &
    \includegraphics[width=0.48\textwidth]{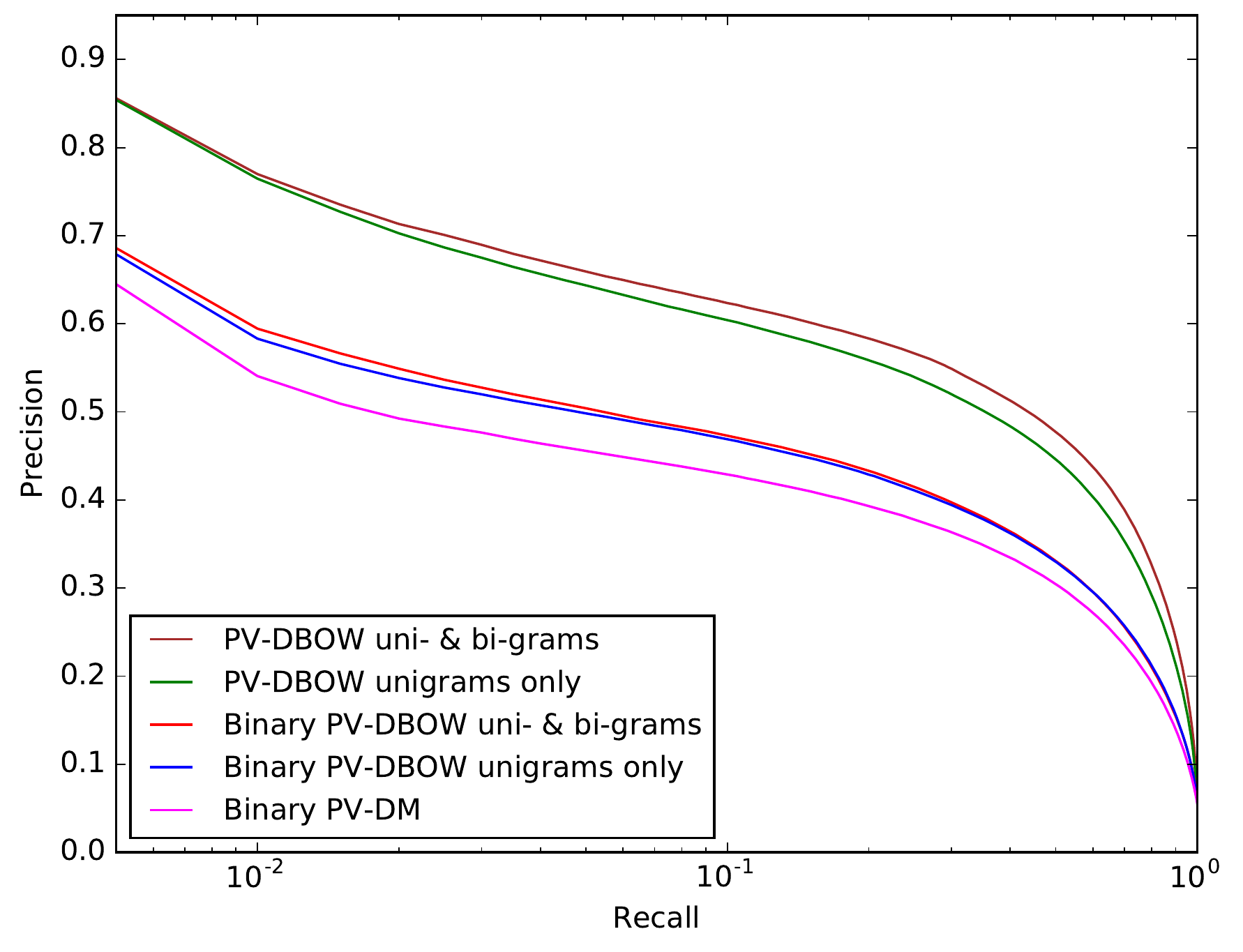} \\
    128 dimensional codes & 32 dimensional codes \vspace{1EM} \\
    \multicolumn{2}{c}{(b) RCV1} \\
    \includegraphics[width=0.48\textwidth]{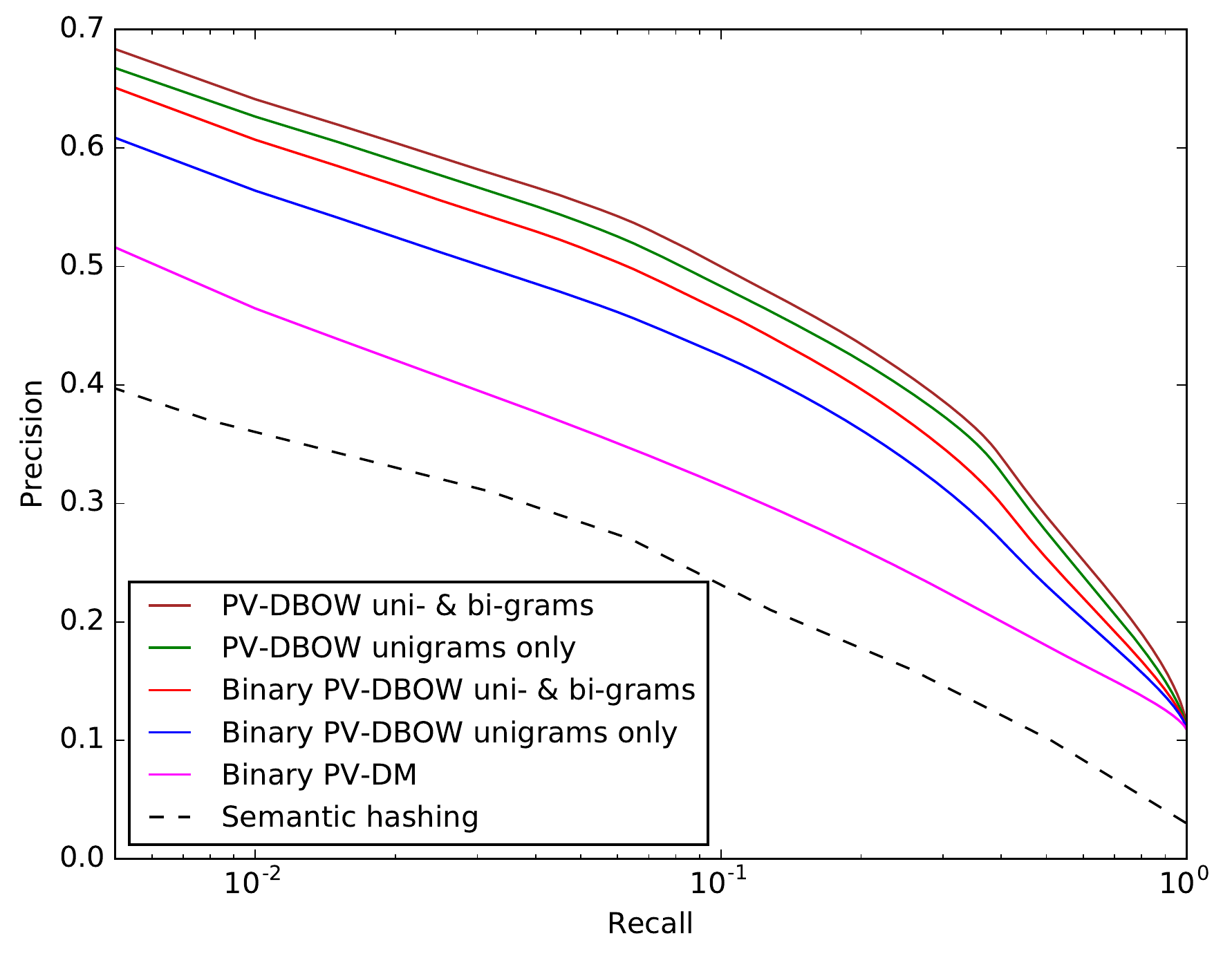} &
    \includegraphics[width=0.48\textwidth]{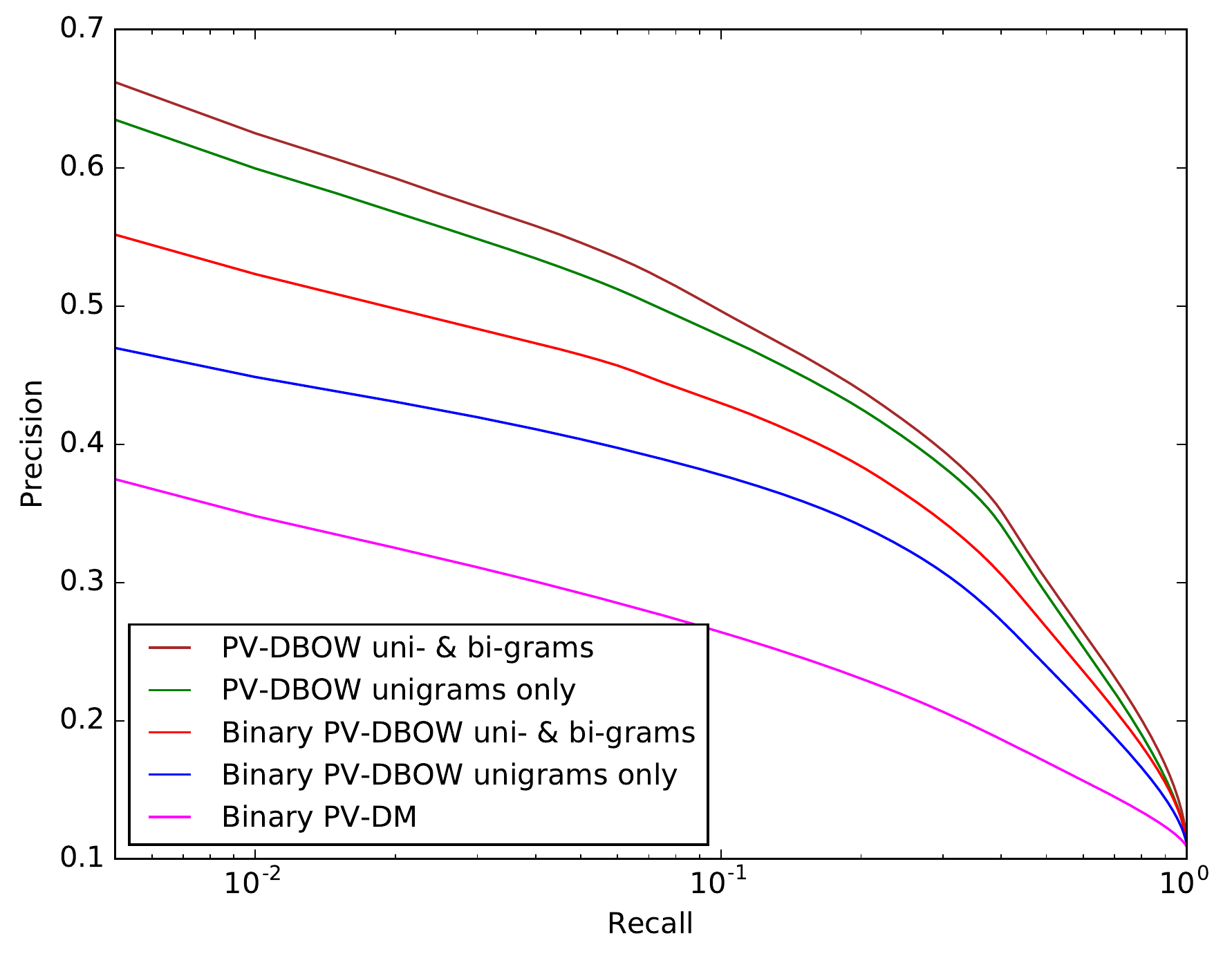} \\
    128 dimensional codes & 32 dimensional codes \vspace{1EM} \\
    \multicolumn{2}{c}{(c) English Wikipedia} \\
    \includegraphics[width=0.48\textwidth]{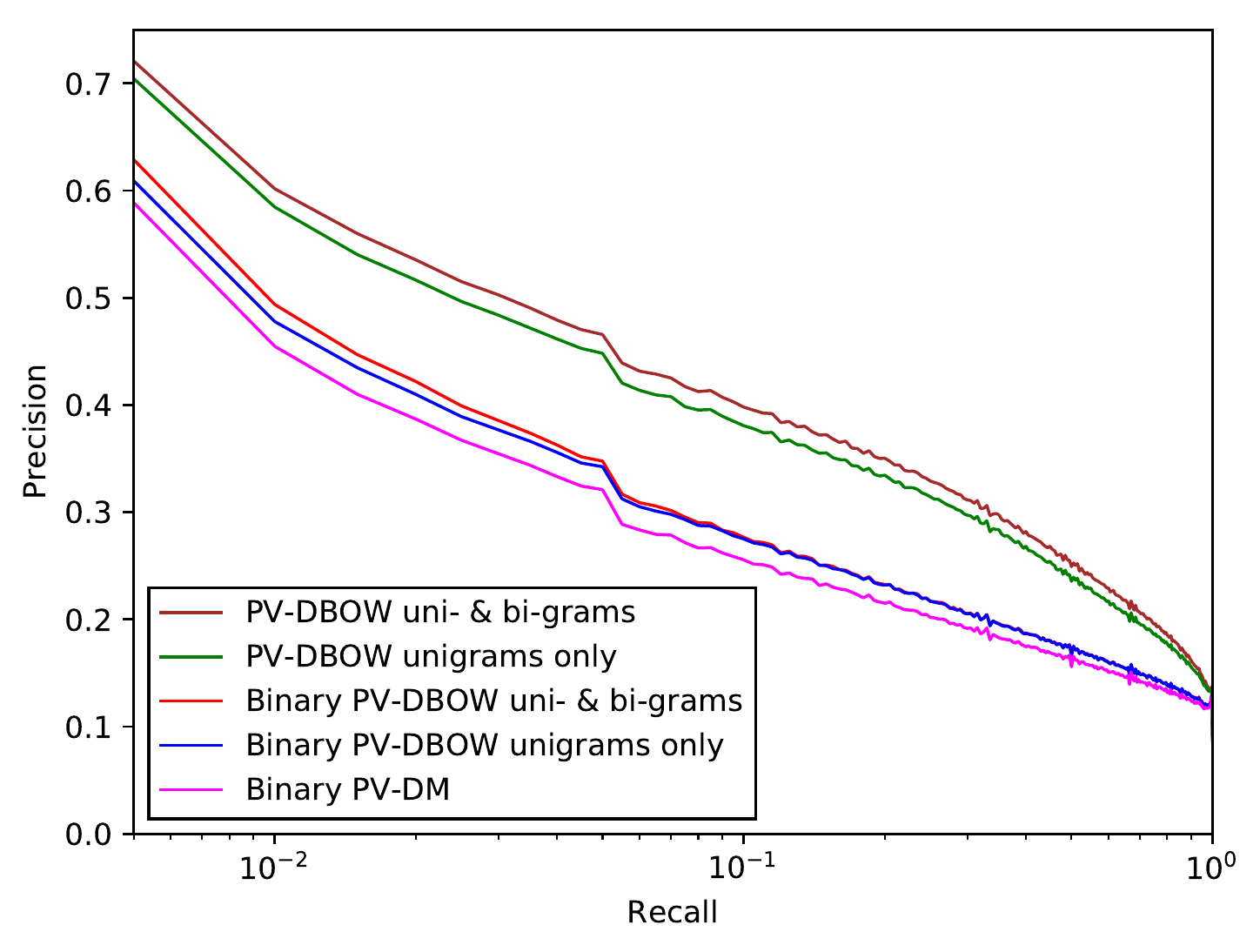} &
    \includegraphics[width=0.48\textwidth]{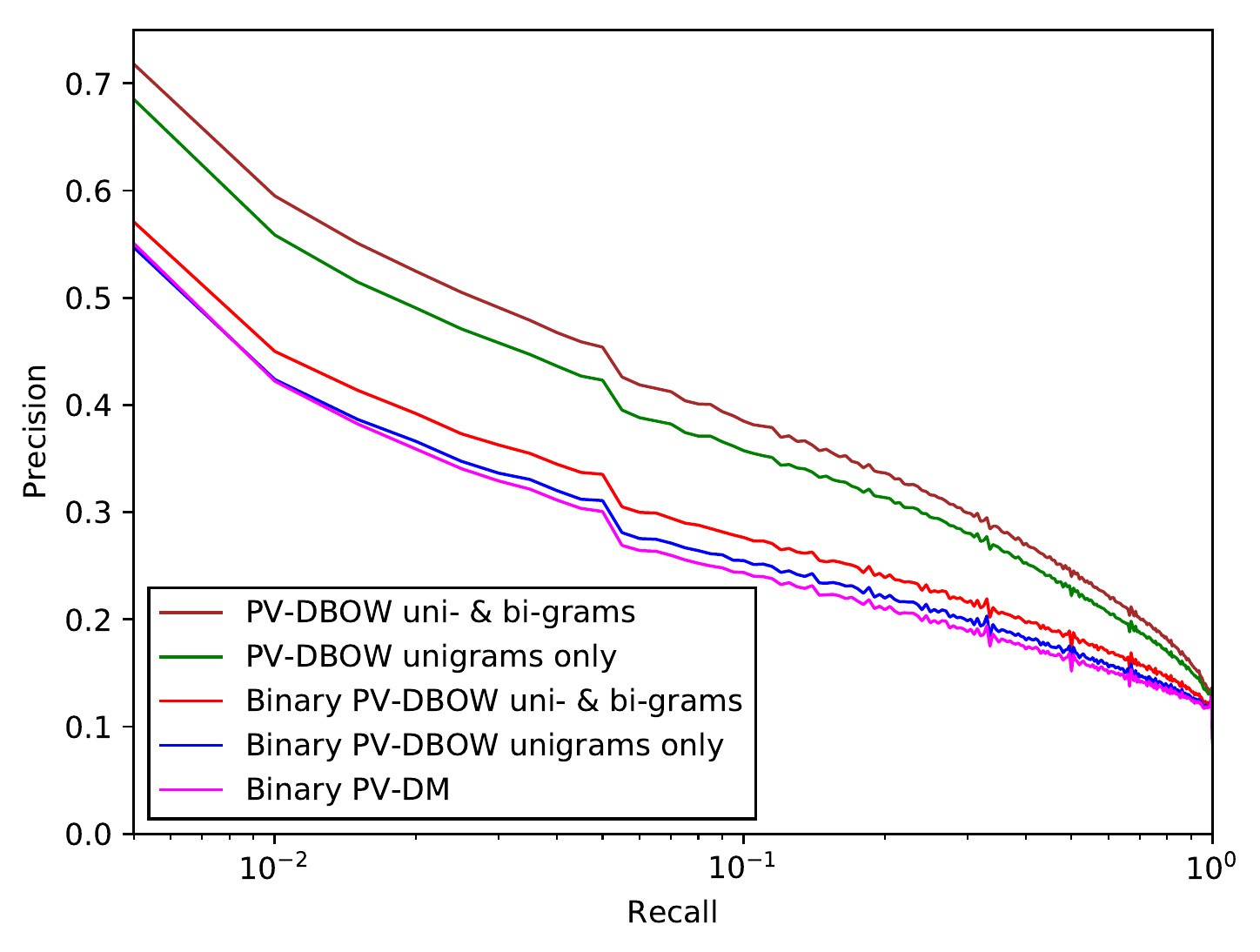} \\
    128 dimensional codes & 32 dimensional codes
  \end{tabular}
  \caption{Precision-recall curves for the 20 Newsgroups, RCV1 and the English Wikipedia. Cosine similarity was used with real-valued
           representations and the Hamming distance with binary codes. For comparison we also included semantic hashing
           results reported by~\citet[Figures~6~\&~7]{salakhutdinov2009semantic}.}
  \label{fig:bpv_precision_recall}
\end{figure*}

\begin{table*}[t]
  \small
  \centering
    \begin{tabular}{|c|c|c|c|c|c|c|c|c|}
      \hline
      Code                 & \multirow{2}{*}{Model}   & With                 & \multicolumn{2}{|c|}{20 Newsgroups} & \multicolumn{2}{|c|}{RCV1}        & \multicolumn{2}{|c|}{English Wikipedia} \\ \cline{4-9}
      size                 &                          & bigrams              & {\small MAP } & {\small NDCG@10 }   & {\small MAP } & {\small NDCG@10 } & {\small MAP } & {\small NDCG@10 }  \\ \hline
      \multirow{7}{*}{128} & \multirow{2}{*}{PV-DBOW} & no                   & 0.4           & 0.75                & 0.25          & 0.79              & 0.25          & 0.59               \\ \cline{3-9}
                           &                          & yes                  & 0.45          & 0.75                & 0.27          & 0.8               & 0.26          & 0.6                \\ \cline{2-9}
                           & Binary                   & no                   & 0.34          & 0.69                & 0.22          & 0.74              & 0.18          & 0.48               \\ \cline{3-9}
                           & PV-DBOW                  & yes                  & \textbf{0.35} & \textbf{0.69}       & \textbf{0.24} & \textbf{0.77}     & \textbf{0.18} & \textbf{0.49}      \\ \cline{2-9}
                           & PV-DM                    & \multirow{2}{*}{N/A} & 0.41          & 0.73                & 0.23          & 0.78              & 0.24          & 0.59               \\ \cline{2-2} \cline{4-9}
                           & Binary PV-DM             &                      & 0.34          & 0.65                & 0.18          & 0.69              & 0.16          & 0.46               \\ \hline
      \multirow{7}{*}{32}  & \multirow{2}{*}{PV-DBOW} & no                   & 0.43          & 0.71                & 0.26          & 0.75              & 0.23          & 0.55               \\ \cline{3-9}
                           &                          & yes                  & 0.46          & 0.72                & 0.27          & 0.77              & 0.25          & 0.58               \\ \cline{2-9}
                           & Binary                   & no                   & 0.32          & 0.53                & 0.22          & 0.6               & 0.16          & 0.41               \\ \cline{3-9}
                           & PV-DBOW                  & yes                  & \textbf{0.32} & \textbf{0.54}       & \textbf{0.25} & \textbf{0.66}     & \textbf{0.17} & \textbf{0.44}      \\ \cline{2-9}
                           & PV-DM                    & \multirow{2}{*}{N/A} & 0.43          & 0.7                 & 0.23          & 0.77              & 0.23          & 0.55               \\ \cline{2-2} \cline{4-9}
                           & Binary PV-DM             &                      & 0.29          & 0.49                & 0.17          & 0.53              & 0.15          & 0.41               \\ \hline
    \end{tabular}
    \caption{Information retrieval results. The best results with binary models are highlighted.}
    \label{tab:bpv_main_results}
\end{table*}

\begin{table*}[b]
  \small
  \centering
    \begin{tabular}{|c|c|c|c|c|c|c|}
      \hline
      \multirow{2}{*}{Hashing algorithm} & \multicolumn{2}{|c|}{20 Newsgroups}           & \multicolumn{2}{|c|}{RCV1}                    & \multicolumn{2}{|c|}{English Wikipedia}       \\ \cline{2-7}
                                         & MAP                   & NDCG@10               & MAP                   & NDCG@10               & MAP                   & NDCG@10               \\ \hline
      Random hyperplane projection       & 0.27                  & 0.53                  & 0.21                  & 0.66                  & 0.16                  & 0.44                  \\ \hline
      Iterative quantization             & 0.31                  & 0.58                  & 0.23                  & 0.68                  & 0.17                  & 0.46                  \\ \hline
    \end{tabular}
    \caption{Information retrieval results for $32$-bit binary codes constructed by first inferring $32$d real-valued paragraph vectors and
             then employing a separate hashing algorithm for binarization. Paragraph vectors were inferred using PV-DBOW with bigrams.}
  \label{tab:bpv_baseline}
\end{table*}

The 20 Newsgroups dataset comes with reference train/test sets. In case of RCV1 we used half of the documents for training and the
other half for evaluation. In case of English Wikipedia we held out for testing randomly selected 10\% of the documents.
We perform document retrieval by selecting queries from the test set and ordering other test documents
according to the similarity of the inferred codes. We use Hamming distance for binary codes and cosine similarity for real-valued
representations. Results are averaged over queries. We assess the performance of our models with precision-recall curves and two
popular information retrieval metrics, namely mean average precision (MAP) and the normalized discounted cumulative gain at
the 10th result (NDCG@10)~\citep{jarvelin2002cumulated}. The results depend, of course, on the chosen document relevancy measure.
Relevancy measure for the 20 Newsgroups dataset is straightforward: a retrieved document is relevant to the query if they both belong
to the same newsgroup. In RCV1 each document belongs to a hierarchy of topics, making the definition of relevancy less
obvious. In this case we adopted the relevancy measure used by~\citet{salakhutdinov2009semantic}. That is, the relevancy is calculated
as the fraction of overlapping labels in a retrieved document and the query document. Overall, our selection of test datasets and
relevancy measures for~20~Newsgroups and RCV1 follows~\citet{salakhutdinov2009semantic}, enabling comparison with semantic hashing codes.
To assess the relevancy of articles in English Wikipedia we can employ categories assigned to them. However, unlike in RCV1, Wikipedia
categories can have multiple parent categories and cyclic dependencies. Therefore, for this dataset we adopted a simplified relevancy measure: two articles are relevant if
they share at least one category. We also removed from the test set categories with less than 20 documents as well as documents that were
left with no categories. Overall, the relevancy is measured over more than $11,800$ categories, making English Wikipedia harder than the
other two benchmarks.

We use AdaGrad~\citep{duchi2011adaptive} for training and inference in all experiments reported in this work. During training we employ
dropout~\citep{srivastava2014dropout} in the embedding layer. To facilitate models with large vocabularies, we approximate the gradients
with respect to the softmax logits using the method described by~\citet{cho2015using}. Binary PV-DM networks use the same number of
dimensions for document codes and word embeddings.

Performance of $128$- and $32$-bit binary paragraph vector codes is reported in~\tabref{bpv_main_results} and in~\figref{bpv_precision_recall}.
For comparison we also report performance of real-valued paragraph vectors. Note that the binary codes perform very well, despite their far lower
capacity: on 20 Newsgroups and RCV1 the $128$-bit Binary PV-DBOW trained with bigrams approaches the performance of the real-valued paragraph
vectors, while on English Wikipedia its performance is slightly lower. Furthermore, Binary PV-DBOW with bigrams outperforms semantic hashing
codes: comparison of precision-recall curves from Figures~\ref{fig:bpv_precision_recall}a and~\ref{fig:bpv_precision_recall}b
with~\citet[Figures 6 \& 7]{salakhutdinov2009semantic} shows that \mbox{$128$-bit} codes learned with this model outperform $128$-bit semantic
hashing codes on 20 Newsgroups and RCV1. Moreover, the $32$-bit codes from this model outperform $128$-bit semantic hashing codes on the RCV1
dataset, and on the 20 Newsgroups dataset give similar precision up to approximately 3\% recall and better precision for higher recall levels.
Note that the difference in this case lies not only in retrieval precision: the short $32$-bit Binary PV-DBOW codes are more efficient for
indexing than long~\mbox{$128$-bit} semantic hashing codes.

We also compared binary paragraph vectors against codes constructed by first inferring short, real-valued paragraph vectors and then using a
separate hashing algorithm for binarization. When the dimensionality of the paragraph vectors is equal to the size of binary codes, the number
of network parameters in this approach is similar to that of Binary PV models. We experimented with two standard hashing algorithms, namely
random hyperplane projection \citep{charikar2002similarity} and iterative quantization \citep{gong2011iterative}. Paragraph vectors in these
experiments were inferred using PV-DBOW with bigrams. Results reported in~\tabref{bpv_baseline} show no benefit from using a separate algorithm
for binarization. On the 20 Newsgroups and RCV1 datasets Binary PV-DBOW yielded higher MAP than the two baseline approaches. On English Wikipedia
iterative quantization achieved MAP equal to Binary PV-DBOW, while random hyperplane projection yielded lower MAP. Some gain in precision of top
hits can be observed for iterative quantization, as indicated by NDCG@10. However, precision of top hits can also be improved by querying with
Real-Binary PV-DBOW model (\sectionref{real-binary-retrieval}). It is also worth noting that end-to-end inference in Binary PV models
is more convenient than inferring real-valued vectors and then using another algorithm for hashing.

\citet{li2015learning} argue that PV-DBOW outperforms PV-DM on a sentiment classification task, and demonstrate that the performance of PV-DBOW
can be improved by including bigrams in the vocabulary. We observed similar results with Binary PV models. That is, including
bigrams in the vocabulary usually improved retrieval precision. Also, codes learned with Binary PV-DBOW provided higher retrieval precision
than Binary PV-DM codes. Furthermore, to choose the context size for the Binary PV-DM models, we evaluated several networks on validation
sets taken out of the training data. The best results were obtained with a minimal one-word, one-sided context window. This is the distributed
memory architecture most similar to the Binary PV-DBOW model.

\subsection{Transfer learning}

\begin{figure}[b!]
  \centering
  \begin{subfigure}[b]{1.0\linewidth}
    \centering
    \includegraphics[width=\textwidth]{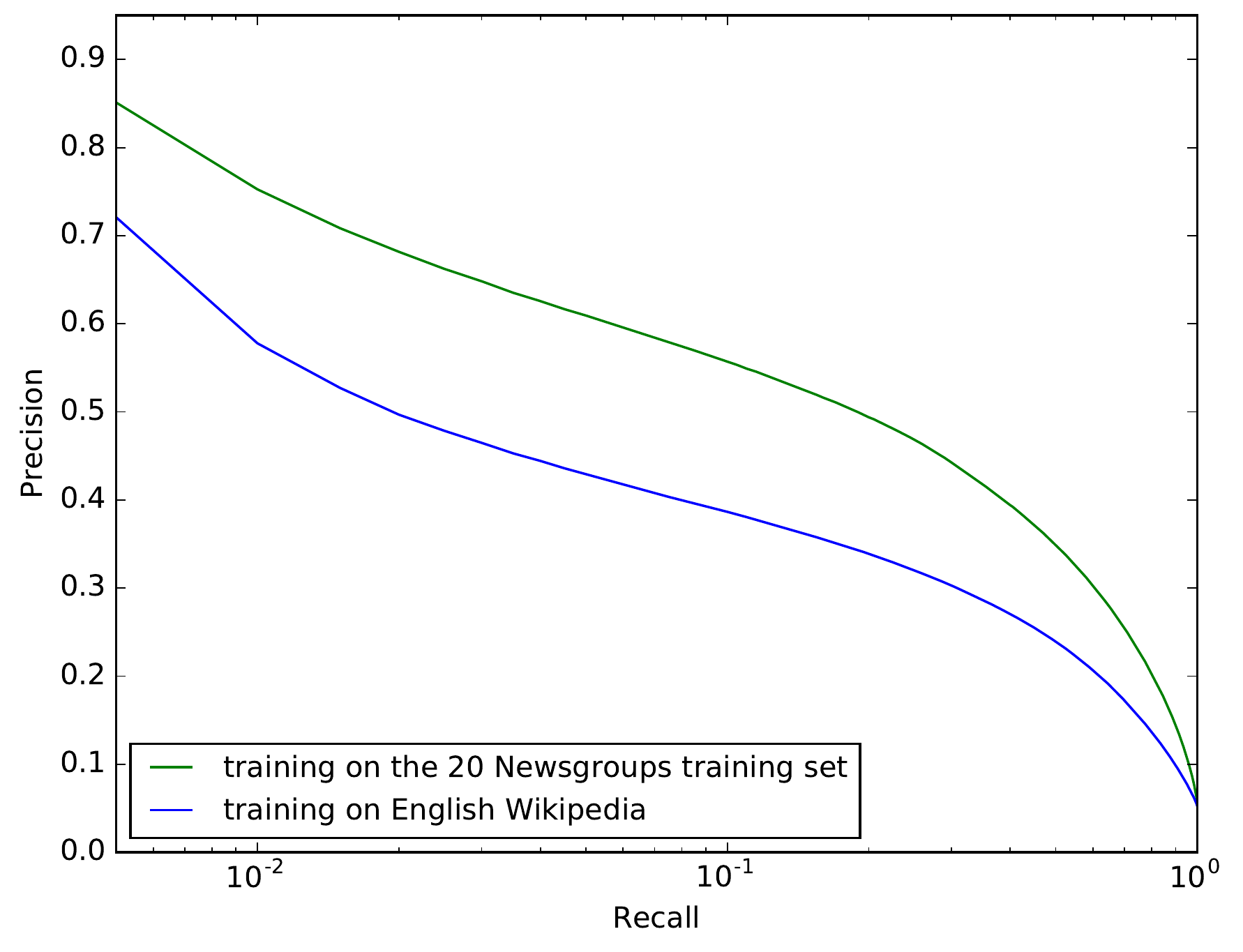}
    \caption{20 Newsgroups}
  \end{subfigure}
  \hfill
  \begin{subfigure}[b]{1.0\linewidth}
    \centering
    \includegraphics[width=\textwidth]{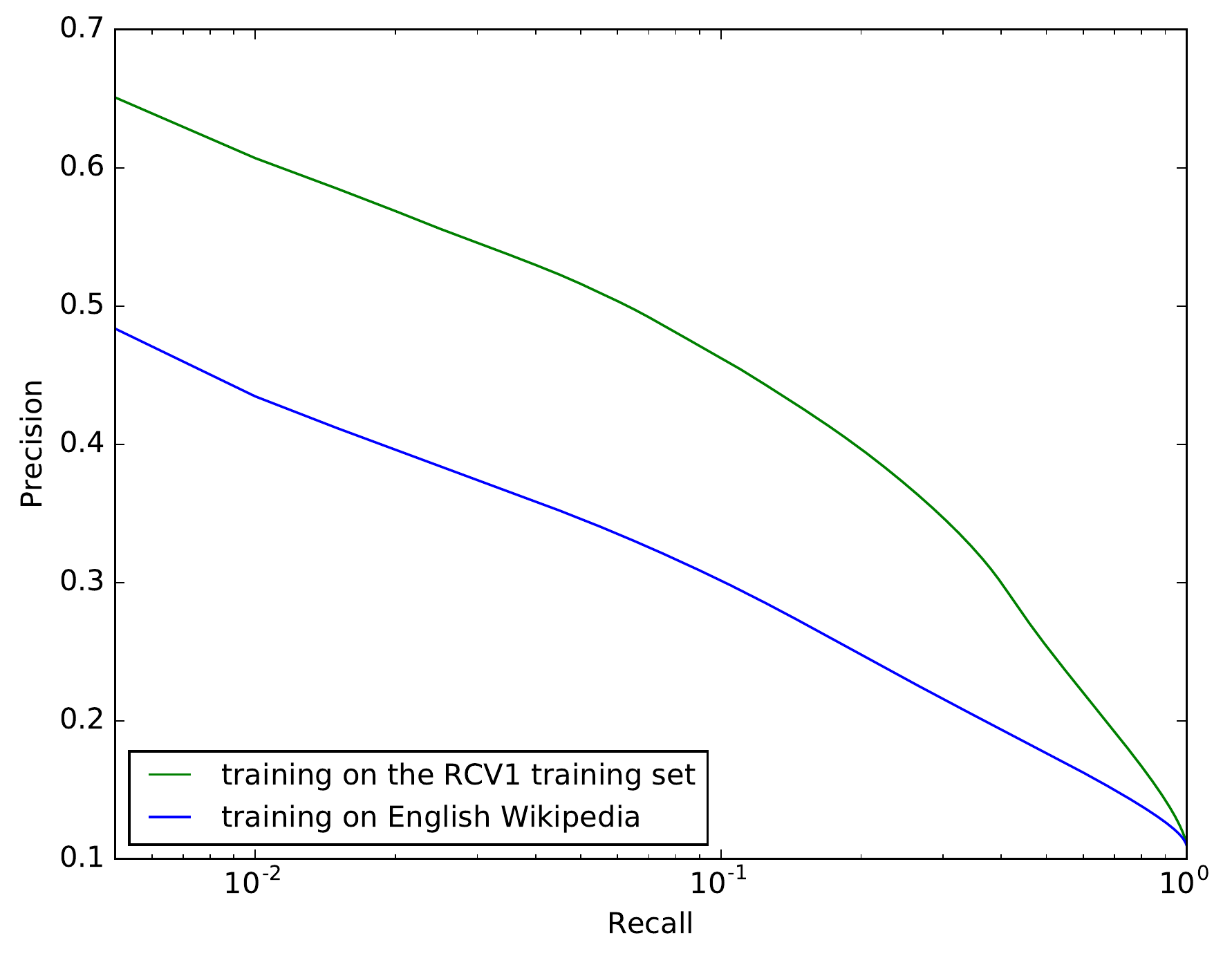}
    \caption{RCV1}
  \end{subfigure}
  \caption{Precision-recall curves for the baseline Binary PV-DBOW models and a Binary PV-DBOW model trained on an unrelated text
           corpus. Results are reported for 128-bit codes.}
  \label{fig:bpv_transfer_learning}
\end{figure}
In the experiments presented thus far we had at our disposal training sets with documents similar to the documents for which we inferred
binary codes. One could ask a question, if it is possible to use binary paragraph vectors without collecting a domain-specific training set?
For example, what if we needed to hash documents that are not associated with any available domain-specific corpus? One solution could be to train
the model with a big generic text corpus, that covers a wide variety of domains. \citet{lau2016empirical} evaluated this approach for real-valued
paragraph vectors, with promising results. It is not obvious, however, whether short binary codes would also perform well in similar settings.
To shed light on this question we trained Binary PV-DBOW with bigrams on the English Wikipedia, and then inferred binary codes for the
test parts of the 20 Newsgroups and RCV1 datasets. The results are presented in~\tabref{bpv_transfer_learning} and in~\figref{bpv_transfer_learning}.
The model trained on an unrelated text corpus gives lower retrieval precision than models with domain-specific
training sets, which is not surprising. However, it still performs remarkably well, indicating that the semantics it captured can be
useful for different text collections. Importantly, these results were obtained without domain-specific finetuning.
\begin{table}[htb]
  \small
  \centering
    \begin{tabular}{|c|c|c|}
      \cline{2-3}
      \multicolumn{1}{c|}{} & MAP   & NDCG@10 \\ \hline
      20 Newsgroups         & 0.24  & 0.51    \\ \hline
      RCV1                  & 0.18  & 0.66    \\ \hline
    \end{tabular}
    \caption{Information retrieval results for the Binary PV-DBOW model trained on an unrelated text corpus. Results are reported
           for~\mbox{128-bit} codes.}
  \label{tab:bpv_transfer_learning}
\end{table}

\subsection{Retrieval with Real-Binary models}\label{sec:real-binary-retrieval}

\begin{figure*}[t!]
  \centering
  \begin{subfigure}[b]{0.49\linewidth}
    \centering
    \includegraphics[width=\textwidth]{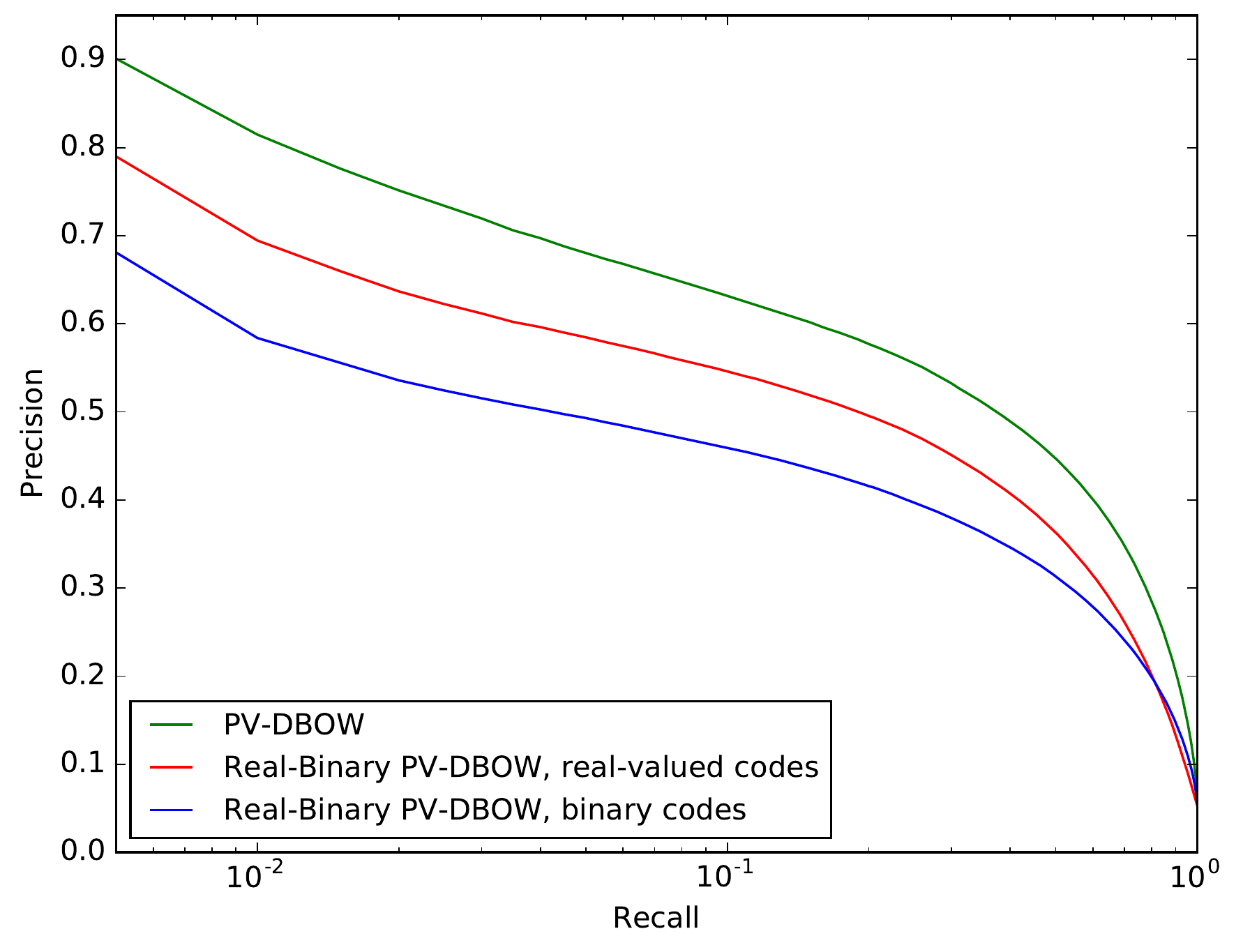}
    \caption{20 Newsgroups}
  \end{subfigure}
  \hfill
  \begin{subfigure}[b]{0.49\linewidth}
    \centering
    \includegraphics[width=\textwidth]{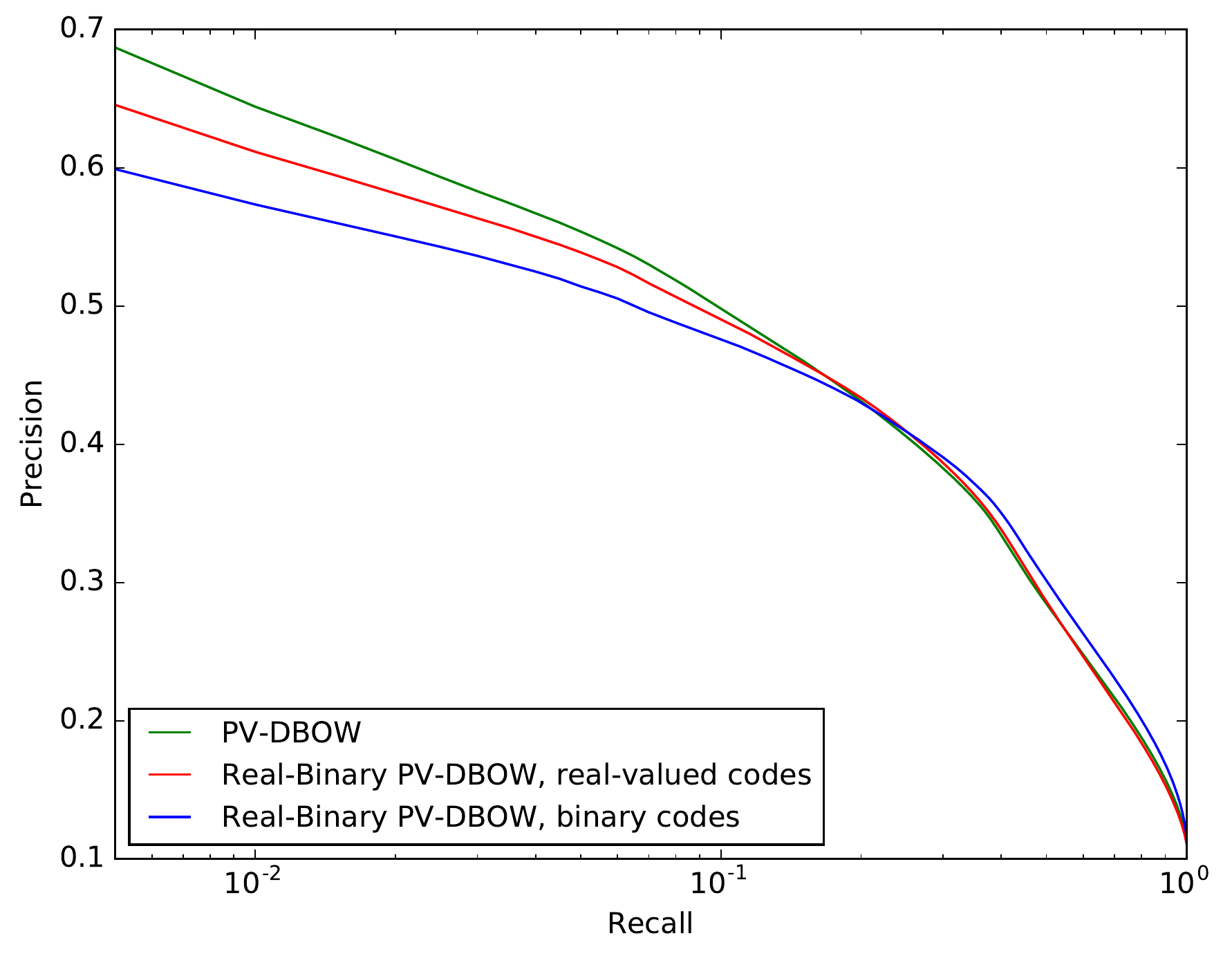}
    \caption{RCV1}
  \end{subfigure}
  \caption{Information retrieval results for binary and real-valued codes learned by the Real-Binary PV-DBOW model with bigrams. Results
           are reported for $28$-bit binary codes and $300$d real-valued codes. A~$300$d PV-DBOW model is included for reference.}
  \label{fig:bpv_real_binary_compare}
\end{figure*}

As pointed out by~\citet{salakhutdinov2009semantic}, when working with large text collections one can use short binary codes for indexing
and a representation with more capacity for ranking. Following this idea, we proposed Real-Binary PV-DBOW model (\sectionref{models}) that
can simultaneously learn short binary codes and high-dimensional real-valued representations. We begin evaluation of this model by comparing
retrieval precision of real-valued and binary representations learned by it. To this end, we trained a Real-Binary PV-DBOW model with $28$-bit
binary codes and $300$-dimensional real-valued representations on the 20 Newsgroups and RCV1 datasets. Results are reported
in~\figref{bpv_real_binary_compare}. The real-valued representations learned with this model give lower precision than PV-DBOW vectors but,
importantly, improve precision over binary codes for top ranked documents. This justifies their use alongside binary codes.

\begin{table}[!b]
  \small
  \centering
    \begin{tabular}{|c|c|c|c|c|c|c|c|}
      \hline
      \multirow{3}{*}{\rot{Code size}} & \multirow{3}{*}{\rot{Radius}} & \multicolumn{6}{c|}{NDCG@10}                                                            \\ \cline{3-8}
                                       &                               & \multicolumn{2}{c|}{20 NG} & \multicolumn{2}{c|}{RCV1} & \multicolumn{2}{c|}{Wikipedia} \\ \cline{3-8}
                                       &                               & A    & B                   & A    & B                  & A    & B                       \\ \hline
      \multirow{2}{*}{28}              & 1                             & 0.79 & 0.85                & 0.77 & 0.85               & 0.66 & 0.7                     \\ \cline{2-1} \cline{3-8}
                                       & \multirow{2}{*}{2}            & 0.72 & 0.8                 & 0.73 & 0.81               & 0.62 & 0.65                    \\ \cline{1-1} \cline{3-8}
      \multirow{2}{*}{24}              &                               & 0.65 & 0.79                & 0.7  & 0.76               & 0.56 & 0.59                    \\ \cline{2-2} \cline{3-8}
                                       & 3                             & 0.63 & 0.76                & 0.69 & 0.74               & 0.5  & 0.55                    \\ \hline
    \end{tabular}
    \caption{Information retrieval results for the Real-Binary PV-DBOW model. Real-valued representations have $300$ dimensions.
             (A) Binary codes are used for selecting documents within a given Hamming distance to the query and real-valued
             representations are used for ranking. (B) For comparison, variant~A was repeated with binary codes inferred using
             plain Binary PV-DBOW and real-valued representation inferred using original PV-DBOW model.}
  \label{tab:bpv_real_binary_performance}
\end{table}

Using short binary codes for initial filtering of documents comes with a tradeoff between the retrieval performance and the recall level.
For example, one can select a small subset of similar documents by using 28--32 bit codes and retrieving documents within small Hamming
distance to the query. This will improve retrieval performance, and possibly also precision, at the cost of recall. Conversely, short codes
provide a less fine-grained hashing and can be used to index documents within larger Hamming distance to the query. They can therefore be
used to improve recall at the cost of retrieval performance, and possibly also precision. For these reasons, we evaluated
Real-Binary PV-DBOW models with different code sizes and under different limits on the Hamming distance to the query. In general, we cannot
expect these models to achieve 100\% recall under the test settings. Furthermore, recall will vary on query-by-query basis. We therefore
decided to focus on the NDCG@10 metric in this evaluation, as it is suited for measuring model performance when a short list of relevant
documents is sought, and the recall level is not known. MAP and precision-recall curves are not applicable in these settings.

Information retrieval results for Real-Binary PV-DBOW are summarized in~\tabref{bpv_real_binary_performance}. The model gives
higher NDCG@10 than $32$-bit Binary PV-DBOW codes (\tabref{bpv_main_results}). The difference is large when the initial filtering is
restrictive, e.g. when using $28$-bit codes and $1$-$2$ bit Hamming distance limit. Real-Binary PV-DBOW can therefore be useful when one needs
to quickly find a short list of relevant documents in a large text collection, and the recall level is not of primary importance. If needed,
precision can be further improved by using plain Binary PV-DBOW codes for filtering and standard DBOW representation for raking
(\tabref{bpv_real_binary_performance}, column~B). Note, however, that PV-DBOW model would then use approximately~$10$~times more
parameters than Real-Binary PV-DBOW.

\section{Conclusion}

In this article we presented simple neural networks that learn short binary codes for text documents. Our networks extend Paragraph Vector
by introducing a sigmoid nonlinearity before the softmax that predicts words in documents. Binary codes inferred with the proposed networks
achieve higher retrieval precision than semantic hashing codes on two popular information retrieval benchmarks. They also retain a lot of
their precision when trained on an unrelated text corpus. Finally, we presented a network that simultaneously learns short binary codes and
longer, real-valued representations.

The best codes in our experiments were inferred with Binary PV-DBOW networks. The Binary PV-DM model did not perform so well.
\citet{li2015learning} made similar observations for Paragraph Vector models, and argue that in distributed memory model the word context
takes a lot of the burden of predicting the central word from the document code. An interesting line of future research could, therefore,
focus on models that account for word order, while learning good binary codes. It is also worth noting that \citet{le2014distributed}
constructed paragraph vectors by combining DM and DBOW representations. This strategy may proof useful also with binary codes, when employed
with hashing algorithms designed for longer codes, e.g. with multi-index hashing~\citep{norouzi2012fast}.

\section*{Acknowledgments}

This research is supported by National Science Centre, Poland grant no.~\mbox{2013/09/B/ST6/01549}
``Interactive Visual Text Analytics~(IVTA): Development of novel, user-driven text mining
and visualization methods for large text corpora exploration.'' This research was carried out with the support of
the ``HPC Infrastructure for Grand Challenges of Science and Engineering'' project, co-financed by the European
Regional Development Fund under the Innovative Economy Operational Programme. This research was supported in part
by PL-Grid Infrastructure.

\appendix

\section{Visualization of Binary PV codes}

\begin{figure}[!t]
  \centering
  \begin{subfigure}[b]{1.0\linewidth}
    \centering
    \includegraphics[width=\textwidth]{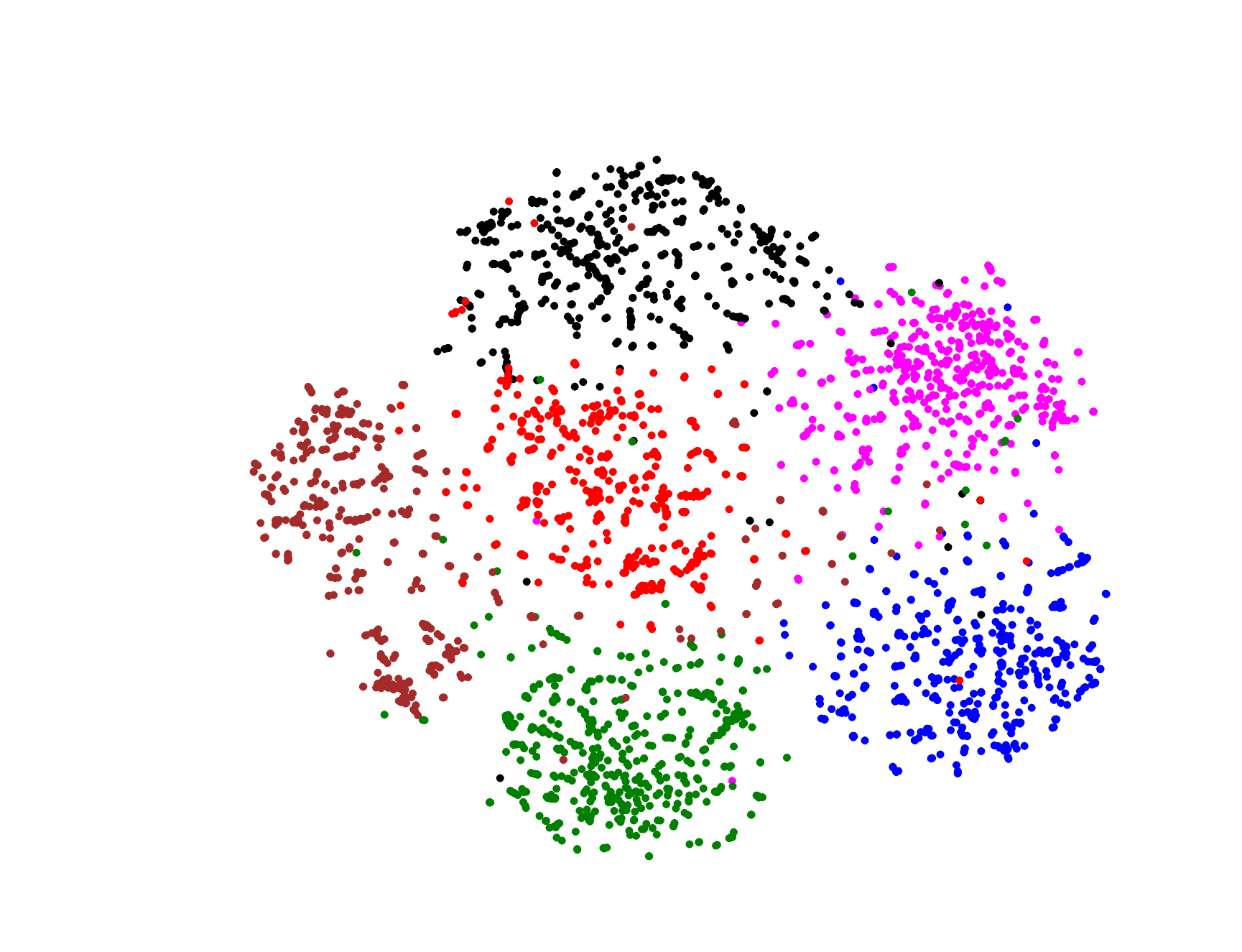}
    \caption{A subset of the 20 Newsgroups dataset: green - soc.religion.christian, red - talk.politics.guns,
                        blue - rec.sport.hockey, brown - talk.politics.mideast, magenta - comp.graphics, black - sci.crypt.}
  \end{subfigure}
  \begin{subfigure}[b]{1.0\linewidth}
    \centering
    \includegraphics[width=\textwidth]{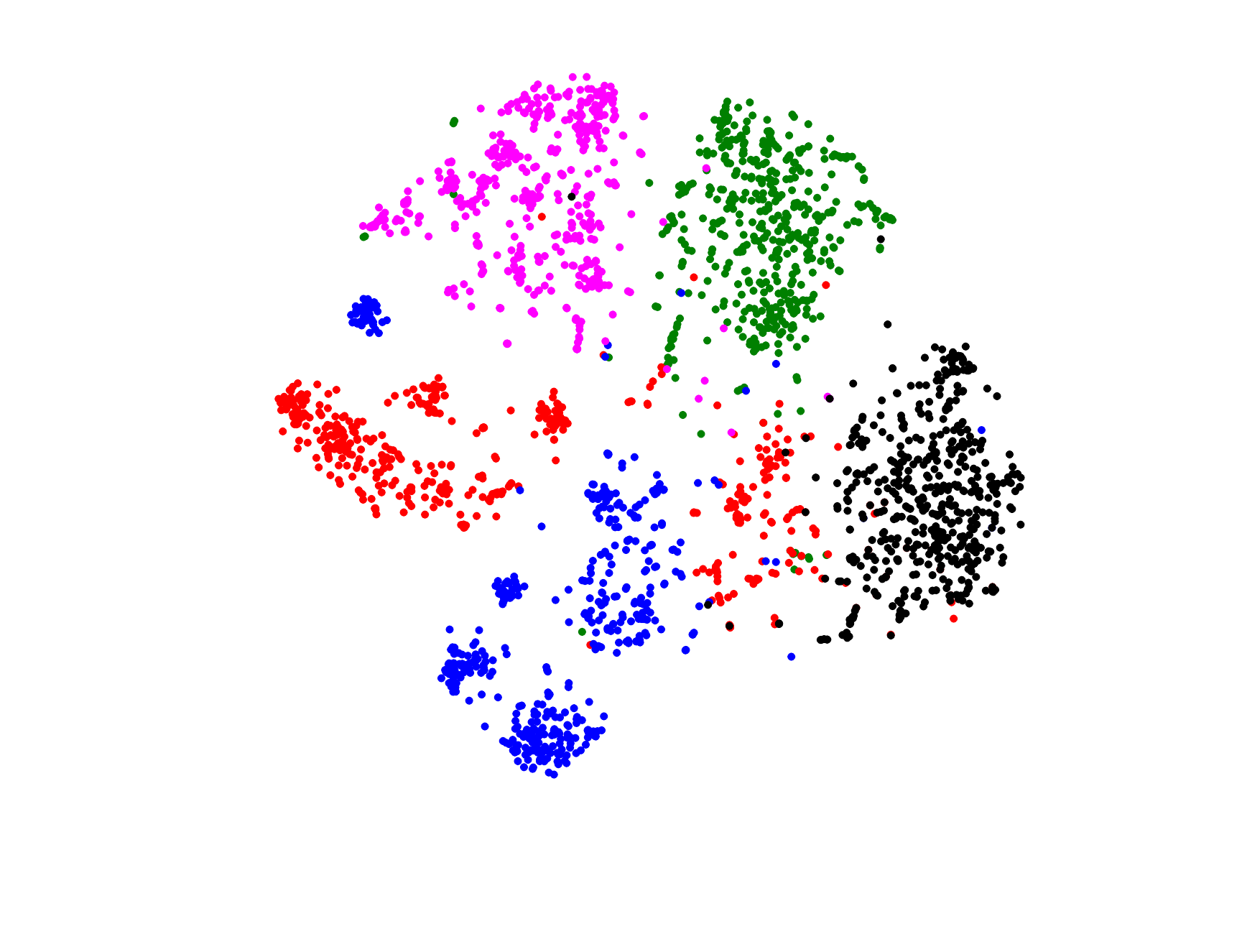}
    \caption{A subset of the RCV1 dataset: green - disasters and accidents, red - government borrowing,
                       blue - accounts/earnings, magenta - energy markets, black - EC monetary/economic.}
  \end{subfigure}
  \caption{t-SNE visualizations of 128 dimensional binary paragraph vector codes; the Hamming distance was used to
           calculate code similarity.}
  \label{fig:binpv_tsne}
\end{figure}

For an additional comparison with semantic hashing, we used t-distributed Stochastic Neighbor Embedding~\citep{maaten2008visualizing} to
construct two-dimensional visualizations of codes learned by Binary PV-DBOW with bigrams. We used the same subsets of newsgroups
and RCV1 topics that were used by~\citet[Figure 5]{salakhutdinov2009semantic}. Codes learned by Binary PV-DBOW (\figref{binpv_tsne})
appear slightly more clustered.

~

\bibliography{bibliography}

\end{document}